\pdfoutput=1

\documentclass[11pt]{article}

\usepackage[]{EMNLP2022}

\usepackage{times}
\usepackage{latexsym}

\usepackage[T1]{fontenc}

\usepackage[utf8]{inputenc}

\usepackage{microtype}

\usepackage{graphicx}
\usepackage{multirow}
\usepackage{makecell}
\usepackage{hyperref}
\usepackage{amsmath,amssymb}
\usepackage{adjustbox}
\usepackage{fdsymbol}
\newcommand{\feta}{\textsc{FETA}}
\definecolor{donkey_brown}{HTML}{A28973}
\definecolor{azure}{HTML}{355AA3}

\usepackage{amsthm}
\newtheorem{definition}{Definition}

\title{\feta: A Benchmark for Few-Sample Task Transfer\\ in Open-Domain Dialogue}

\author{Alon Albalak\textsuperscript{1}\quad
\textbf{Yi-Lin Tuan\textsuperscript{1}\quad Pegah Jandaghi\textsuperscript{2}\quad Connor Pryor\textsuperscript{3}\quad Luke Yoffe\textsuperscript{1}}\\
\bf{Deepak Ramachandran\textsuperscript{4}\quad Lise Getoor\textsuperscript{3}\quad Jay Pujara\textsuperscript{2}\quad William Yang Wang\textsuperscript{1}}\\
\textsuperscript{1}University of California, Santa Barbara\quad
\textsuperscript{2}University of Southern California\\
\textsuperscript{3}University of California, Santa Cruz\quad
\textsuperscript{4}Google Research\\
\texttt{alon\_albalak@ucsb.edu}\\}

\begin{document}
\maketitle

\begin{abstract}
Task transfer, transferring knowledge contained in related tasks, holds the promise of reducing the quantity of labeled data required to fine-tune language models. Dialogue understanding encompasses many diverse tasks, yet task transfer has not been thoroughly studied in conversational AI.
This work explores conversational task transfer by introducing \feta: a benchmark for \textbf{FE}w-sample \textbf{TA}sk transfer in open-domain dialogue.
\feta\;contains two underlying sets of conversations upon which there are 10 and 7 tasks annotated, enabling the study of intra-dataset task transfer; task transfer without domain adaptation.
We utilize three popular language models and three learning algorithms to analyze the transferability between 132 source-target task pairs and create a baseline for future work.
We run experiments in the single- and multi-source settings and report valuable findings, e.g., most performance trends are model-specific, and span extraction and multiple-choice tasks benefit the most from task transfer.
In addition to task transfer, \feta\;can be a valuable resource for future research into the efficiency and generalizability of pre-training datasets and model architectures, as well as for learning settings such as continual and multitask learning.
\footnote{Benchmark available at \href{https://alon-albalak.github.io/feta-website/}{alon-albalak.github.io/feta-website}. We utilize the Transfer Learning in Dialogue Benchmarking Toolkit for all experiments (\href{https://github.com/alon-albalak/TLiDB}{TLiDB python package}).}
\end{abstract}
\section{Introduction}
\label{sec:intro}

\begin{figure}[t!]
    \centering
    \includegraphics[width=\columnwidth]{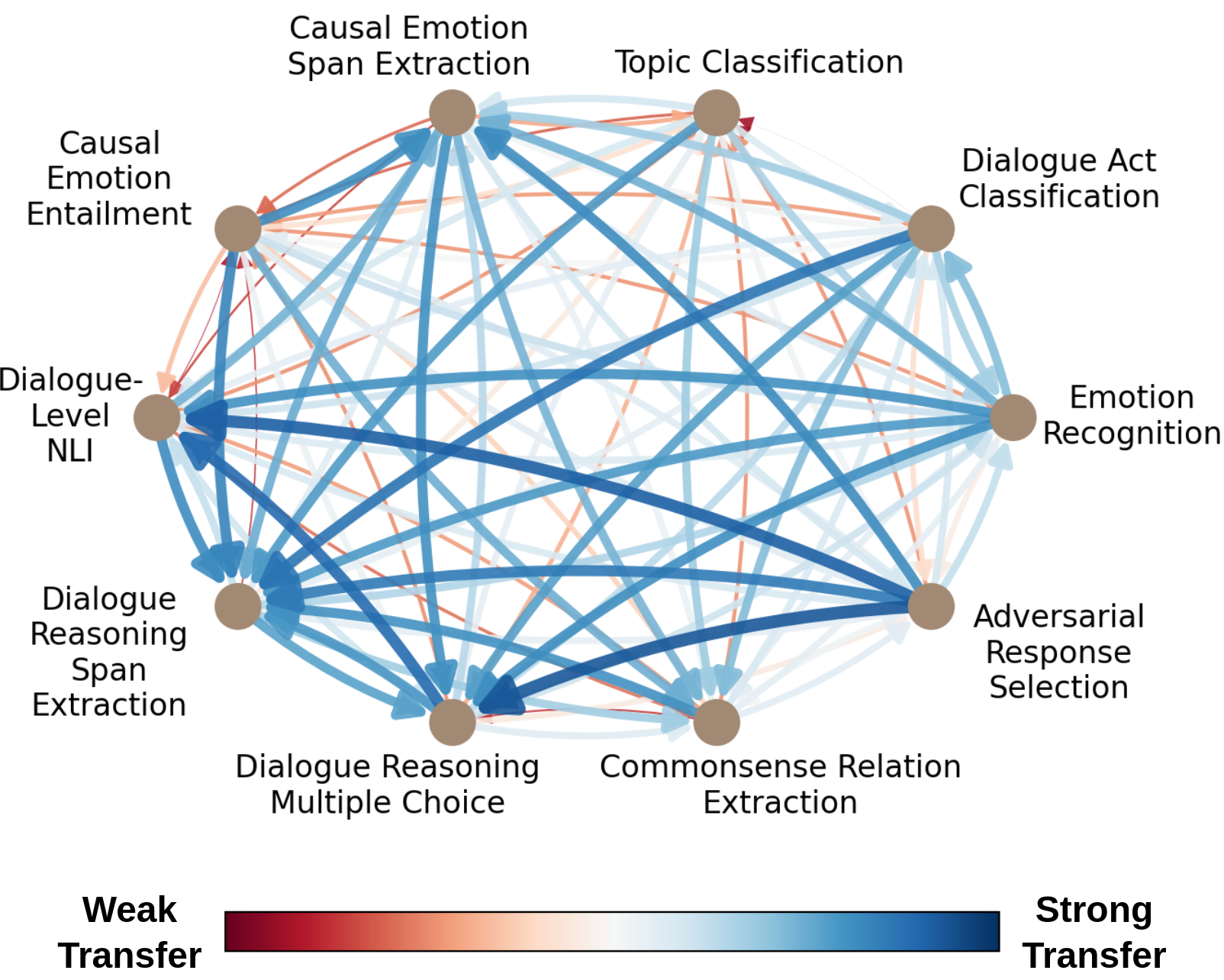}
    \caption{\textbf{Task Transfer Performance} on \feta-DailyDialog. Computed transfer performance is demonstrated by arrows leaving from source tasks and entering target tasks. Strength of the transfer is denoted by thickness and color of edges.}
    \label{fig:circles}
\end{figure}

Improving sample efficiency through transfer learning has been a long-standing challenge in the machine learning and natural language processing communities \cite{transfer_91,JMLR:v6:ando05a}.
Dialogue data requires multiple cohesive turns with consistent speaker personalities \cite{urbanek-etal-2019-learning,Huang2020ChallengesIB}, creating a challenge for data collection and motivating the development of techniques that improve sample efficiency in conversational AI \cite{lin-etal-2020-mintl}.

Furthermore, dialogue understanding tasks require a shared knowledge of semantics, pragmatics, human behavior, and commonsense, making dialogue an area of study that can benefit greatly from a deeper understanding of transfer learning.

Two essential transfer learning settings, namely domain adaptation and task transfer, have been studied on language tasks \cite{ruder-etal-2019-transfer}. While domain adaptation has been studied in task-oriented dialogue \cite{MehriDialoGLUE2020}
, task transfer has been studied with less rigor in conversational AI. Prior studies of task transfer in dialogue consider only 2-4 tasks, focus on multitask learning, and do not compare learning algorithms \cite{simpleTOD,10.1162/tacl_a_00399}.

Prior studies have focused on cross-dataset task transfer, gathering tasks annotated on disjoint datasets \cite{Vu2020ExploringAP, Ye2021CrossFitAF}, but this can lead to improvements in domain adaptation being confounded as improvements in task transfer. A precise study of task transfer should be on a single data source in an intra-dataset transfer setting, as in \citet{Zamir_2018_CVPR}. Additionally, previous studies focus on learning algorithms and use only a single language model architecture \cite{pruksachatkun-etal-2020-intermediate,Lourie2021UNICORNOR,aribandi2022ext}, which may lead to a narrow understanding.
To the best of our knowledge, this is the first rigorous study on task transfer in dialogue and the most extensive intra-dataset task transfer study in NLP.

In this work, we create \feta, a benchmark for few-sample task transfer for language understanding in open-domain dialogue with 17 total tasks. \feta\;datasets cover a variety of properties (dyadic vs. multi-party, anonymized vs. recurring speaker, varying dialogue lengths) and task types (utterance-level classification, dialogue-level classification, span extraction, multiple-choice), and maintain a wide variety of data quantities.

We study task transfer on \feta\; by comparing three task transfer algorithms and three commonly used language models in single-source and multi-source settings. Figure \ref{fig:circles} illustrates some results in the single-source setting. For example, we find that Dialogue Reasoning Span Extraction benefits from nearly all source tasks. On the other hand, Adversarial Response Selection and Emotion Recognition improve the performance of many target tasks when utilized as a source task.

In this study, we find that:
\textbf{(i)} Trends are largely model-dependent, a finding that previous works have not discussed.
\textbf{(ii)} Out of all task types, span extraction tasks gain the most as a target, especially with few samples.
\textbf{(iii)} Adding source tasks does not uniformly improve over a single source task,
motivating a better understanding of the complex relationship between source and target tasks.

\feta\;provides a resource for various future studies, e.g., on the generalizability of model architectures, and pre-training datasets that enable efficient transfer. In addition to task transfer, \feta\;can also facilitate the study of continual and multitask learning.

In summary, our main contributions are:
\begin{itemize}
    \item We create the first large-scale benchmark for task transfer in dialogue, with 132 source-target task pairs.
    \item Extensive experimentation on \feta\;in both the single-source and multi-source settings, and an in-depth analysis comparing models, learning algorithms, sample sizes, and task types, finding new and non-intuitive results.
    \item A readily extensible transfer learning framework\footnote{\href{https://github.com/alon-albalak/TLiDB}{github.com/alon-albalak/TLiDB}} that allows for rapid experimentation and an online leaderboard\footnote{\href{https://alon-albalak.github.io/feta-website/}{alon-albalak.github.io/feta-website/}} to encourage deeper research into task transfer.
\end{itemize}
\section{Related Work}
\label{sec:related_work}


\paragraph{Transfer Learning in NLP}
Prior works on transfer learning in NLP have studied a wide variety of topics, including domain adaptation \cite{36364}, multitask learning \cite{10.1145/1390156.1390177,bingel-sogaard-2017-identifying}, and learning representations of words \cite{brown-etal-1992-class,NIPS2013_9aa42b31,peters-etal-2017-semi,peters-etal-2018-deep}.
More recently, DialoGLUE \cite{MehriDialoGLUE2020} and RADDLE \cite{Peng2021RADDLEAE} study domain adaptation for language understanding tasks in task-oriented dialogue.
\citet{Shuster2020TheDD} focuses on multitasking in dialogue response generation across multiple datasets.
Similar to this work, \citet{pruksachatkun-etal-2020-intermediate} study task transfer, although they study cross-dataset task transfer in general NLP tasks.
\citet{Lourie2021UNICORNOR} also study task transfer, but they focus on the T5 model and a suite of commonsenseQA datasets. 


\paragraph{Task Transfer in Dialogue}
Task transfer has been applied in Task-Oriented Dialogue (TOD) settings but never rigorously studied. For example, \citet{simpleTOD} and \citet{lin-etal-2020-mintl} develop multitask models to perform 2-4 TOD tasks but do not aim to analyze the efficiency of models or learning algorithms for task transfer.


\paragraph{Intra-dataset Task Transfer}
Intra-dataset task transfer has been studied in computer vision applications \cite{Zamir_2018_CVPR,pal2019zero}, but to our best knowledge it has never been studied in NLP.




\section{\feta}
\label{sec:feta}

\begin{table*}
    \centering
    \small
    \begin{tabular}{l l r | r r r | l | l}
        & & \multirow{2}{*}{\begin{tabular}{c}\textbf{Original}\\\textbf{Samples}\end{tabular}}& \multicolumn{3}{c|}{\textbf{FETA Samples}} & \multirow{2}{*}{\begin{tabular}{r}\textbf{Task}\\\textbf{Type}\end{tabular}}& \\
         & \textbf{Task Name} & & \textbf{Train} & \textbf{Dev} & \textbf{Test} & & \textbf{Metrics}\\
         \hline
         \parbox[t]{2mm}{\multirow{10}{*}{\rotatebox[origin=c]{90}{\normalsize{DailyDialog}}}} & Emotion Recognition & 102978 & 7230 & 1269 & 15885 & Utt Cls & M/m-F1 \\
         & Dialogue Act Classification & 102978 & 7230 & 1269 & 15885 & Utt Cls & M/m-F1\\
         & Topic Classification & 13118 & 958 & 161 & 1919 & Dial Cls & M/m-F1\\
         & Causal Emotion Span Extraction & 36324 & 2141 & 169 & 9133 & Span Ex & T-F1,EM\\
         & Causal Emotion Entailment & 36324 & 2141 & 169 & 9133 & Dial Cls & M-F1,Acc\\
         & Dialogue-Level NLI & 5817 & 569 & 52 & 1302 & Dial Cls & M-F1,Acc\\
         & Dialogue Reasoning Span Extraction & 1098 & 123 & 13 & 244 & Span Ex & T-F1,EM\\
         & Dialogue Reasoning Multiple Choice & 2165 & 224 & 26 & 496 & Mult Ch & Acc\\
         & Commonsense Relation Extraction & 4009 & 350 & 38 & 851 & Dial Cl. & M-F1,Acc\\
         & Adversarial Response Selection & 57145 & 3400 & 895 & 10750 & Mult Ch & Acc\\
         \hline
         \parbox[t]{2mm}{\multirow{7}{*}{\rotatebox[origin=c]{90}{\normalsize{Friends}}}} & Emotion Recognition (EmoryNLP) & 12606 & 844 & 207 & 1912 & Utt Cls & m/W-F1\\
         & Reading Comprehension & 13865 & 912 & 181 & 2284 & Mult Ch & Acc\\
         & Character Identification & 50247 & 3593 & 638 & 7803 & Utt Cls & M/m-F1\\
         & Question Answering & 12257 & 819 & 191 & 1937 & Span Ex & T-F1,EM\\
         & Personality Detection & 711 & 54 & 15 & 110 & Dial Cls & Acc\\
         & Relation Extraction & 7636 & 519 & 121 & 1188 & Dial Cls & m-F1\\
         & Emotion Recognition (MELD) & 9140 & 616 & 148 & 1247 & Utt Cls & m/W-F1\\
         \hline
    \end{tabular}
    \caption{\textbf{Overview of \feta\;tasks}. Task types are abbreviated as follows: Utt Cls for utterance-level classification, Dial Cls for dialogue-level classification, Span Ex for span extraction, and Mult Ch for multiple choice. Metrics are abbreviated as follows: M-F1 for macro-F1, m-F1 for micro-F1, T-F1 for token-F1, W-F1 for weighted-F1, EM for exact match and Acc for accuracy.}
    \label{tab:dataset_info}
\end{table*}

In this section, we briefly define \textit{intra-dataset task transfer}, the problem setting of \feta. Then, we introduce \feta, our benchmark for few-sample task transfer in open-domain dialogue. Finally, we define the metrics we use to evaluate models and learning algorithms on \feta.

\subsection{Problem Definitions}
Let a dataset be composed of the instance set, $X$, and $n$ task-specific label sets $Y_{1},Y_{2},\dots,Y_{n}$. In \feta, each instance $x\in X$ is a dialogue.

\begin{definition}[Domain and Task]
A \textnormal{domain} $\mathcal{D}=\{\mathcal{X},P(X)\}$ consists of a feature space $\mathcal{X}$ and a marginal probability distribution $P(X)$. The marginal probabilities are over the instance set $X=\{x_{1},x_{2},\dots,x_{n}\}\in\mathcal{X}$.

A \textnormal{task} $\mathcal{T}=\{\mathcal{Y},f(X)\}$ is composed of a label space $\mathcal{Y}$ and a predictive function, $f:\mathcal{X}\rightarrow\mathcal{Y}$.
\end{definition}


\begin{definition}[Learning Algorithm]
A \textnormal{learning algorithm}, $\mathcal{A}$, is a protocol that determines the method by which the instance set $X$ and task-specific label sets $Y_{1},Y_{2},\dots,Y_{n}$ will be used to train a predictive function, $f$.
\end{definition}

\begin{definition}[Task Transfer]
Given a source task $\mathcal{T}_{S}=\{\mathcal{Y}_{S},f_{S}(X_{S})\}$ and target task $\mathcal{T}_{T}=\{\mathcal{Y}_{T},f_{T}(X_{T})\}$, \textnormal{task transfer} is the use of a learning algorithm, $\mathcal{A}$, to improve the learning of $f_{T}$ by using the knowledge in $\mathcal{T}_{S}$.
\end{definition}

In \textbf{cross-dataset task transfer}, when $X_{S} \neq X_{T}$, we also have $P(X_{S}) \neq P(X_{T})$ and $\mathcal{D}_{S} \neq \mathcal{D}_{T}$; domain shift.

In \textbf{intra-dataset task transfer}, when $X_{S}= X_{T}$, there is no domain shift. This enables the study of the learning algorithm's performance on task transfer, isolated from domain adaptation.

We refer the reader to \citet{Pan2010ASO} and \citet{Zhuang2021ACS} for expanded discussions on transfer learning definitions.

\paragraph{Few-Sample}
Due to the challenge and cost of collecting and annotating data, many real-world applications of NLP techniques are limited by data quantities. For this reason, we focus on the few-sample setting, defined in \feta\;as 10\% of the original instance set. Out of 10\%, 5\%, and 1\%, 10\% was empirically determined to be the smallest percentage that retains labels from all label sets in both the train and development partitions. Given the recent attention focused on NLP applications in low-resource settings \cite{NEURIPS2020_1457c0d6,Bansal2020SelfSupervisedMF,mukherjee2021clues, Ye2021CrossFitAF}, we expect research done in such a low-data setting will lead to insights useful for many researchers and practitioners.

\begin{figure*}[t]
    \centering
    \includegraphics[width=0.98\linewidth]{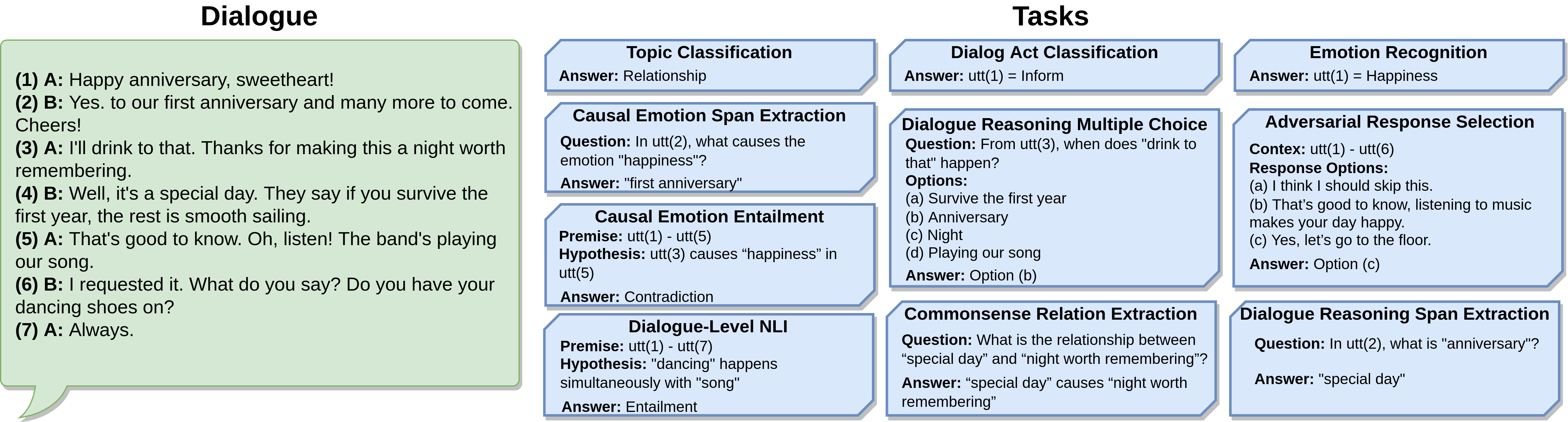}
\end{figure*}


\begin{figure*}[t]
    \centering
    \includegraphics[width=0.98\linewidth]{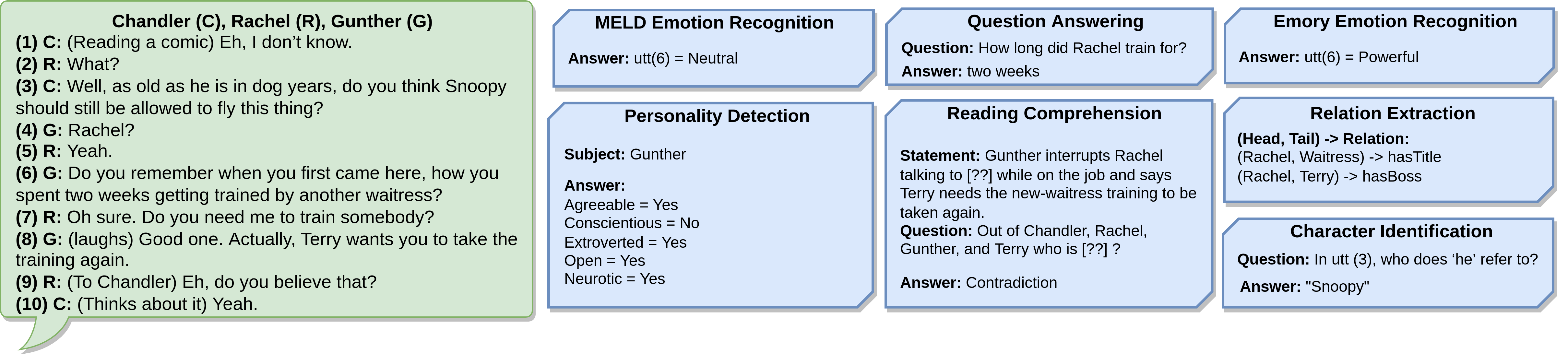}
    \caption{\textbf{Example dialogues and tasks} for FETA-DailyDialog (top) and FETA-Friends (bottom).}
    \label{fig:feta-dialogues}
\end{figure*}

\subsection{\feta\;Datasets}
\label{sec:datasets}
In this section, we describe the two dialogue sources we use, DailyDialog \cite{li-etal-2017-dailydialog} and Friends \cite{chen-choi-2016-character}, and the tasks annotated on each source.


We select these datasets because they complement each other in desirable ways.
DailyDialog contains 2-speaker dialogues where speakers are anonymized and averages 88 words per dialogue. In contrast, Friends consists of multiparty dialogues (3.6 speakers mean, 15 max) with recurring characters and averages 283 words per dialogue.
These differences lead to each set of dialogue instances having different task annotations, giving \feta\;a wider variety of tasks. For example, DailyDialog tasks include understanding the causes of emotions and commonsense reasoning, while tasks annotated on Friends revolve more around recognizing entities and understanding personalities.

To create \feta\;versions of each dataset, we first partition the dialogues into 70/15/15\% splits for training, validation, and test sets. After splitting, we randomly down-sample the train and development dialogues to 10\% of the original quantities. Thus, \feta\;splits use 7/1.5/15\% of the original dialogues. Not every dialogue is annotated for all tasks, allowing some tasks to have more samples than others. Crucially, the data splits are the same for all tasks, preventing data leakage. Table \ref{tab:dataset_info} shows an overview of the tasks, samples, and metrics used for each dataset.

\paragraph{FETA-DailyDialog}
\citet{li-etal-2017-dailydialog} present the DailyDialog dataset, with chit-chat conversations covering 10 various topics including relationships, politics, and work.

Many works add annotations on top of these dialogues and \feta\;utilizes 10 of them.
Figure \ref{fig:feta-dialogues} provides an overview of the tasks: \textit{emotion recognition}, \textit{dialogue act classification}, \textit{topic classification} (from DailyDialog~\cite{li-etal-2017-dailydialog}), \textit{causal emotion span extraction}, \textit{causal emotion entailment} (from RECCON ~\cite{reccon}), \textit{dialogue-level natural language inference}, \textit{dialogue reasoning span extraction}, \textit{dialogue reasoning multiple choice}, \textit{commonsense relation extraction} (from CIDER~\cite{ghosal-etal-2021-cider})
\textit{adversarial response selection} (from DailyDialog++~\cite{dailydialog_plusplus}).
For further details of these tasks, we refer the reader to Appendix~\ref{sec:dataset_details} and their original papers.

\paragraph{FETA-Friends}
The Friends dialogues come from transcripts of 10 seasons of the TV show by the same name \cite{chen-choi-2016-character}. In addition to dialogue, the transcripts contain situational information such as behaviors and non-verbal information like scene information. 

In total, \feta\;has 7 task annotations on top of the Friends scripts. As illustrated in Figure~\ref{fig:feta-dialogues}, the incorporated tasks include \textit{Emory emotion recognition} (from \cite{zahiri2018emotion}), \textit{reading comprehension} (from \cite{ma-etal-2018-challenging}),
\textit{character identification} (from \cite{chen-choi-2016-character, zhou-choi-2018-exist}),
\textit{question answering} (from \cite{yang-choi-2019-friendsqa}),
\textit{personality detection} (from \cite{jiang2020automatic}), and \textit{relation extraction} (from DialogRE \cite{yu2020dialogue}) and \textit{MELD emotion recognition} (from MELD \cite{poria2019meld}). There are two emotion recognition label sets (Emory and MELD), but they have only 22\% overlap in instance sets and have different label spaces. For further details of these tasks, we refer the reader to Appendix~\ref{sec:dataset_details} and their original papers.

\subsection{Evaluation Metrics}
To define the metrics, we consider 4 variables: source task $s$, target task $t$, model $f$, and learning algorithm $\mathcal{A}$, and we abuse notation slightly to allow for $f_\mathcal{A}(s,t)$ to represent a model trained on the source and target tasks using the given learning algorithm. In \feta, we evaluate the performance of a model and learning algorithm with multiple metrics: average and top-1 raw scores, as well as average and top-1 score $\Delta$s.

\paragraph{Average and Top-1 Scores}
First, we consider the two raw scores: average score and top-1 score. These metrics aim to answer the following questions: How well do a model and algorithm perform across all task pairs, and, how well do a model and algorithm perform supposing that we knew the best source task a priori.

We calculate an average score across all source-target task pairs to understand how each model and algorithm performs in the aggregate. Formally, let the score for a single task be computed as:
\[
    score(s,t,f,\mathcal{A}) = \frac{1}{|M_{t}|}\sum\limits_{i=1}^{|M_{t}|}M_{t,i}(f_{\mathcal{A}}(s,t))
\]
where $M_{t}$ is the set of metrics associated with task $t$, found in Table \ref{tab:dataset_info}, and $M_{t,i}(f)$ is the $i$th calculated metric of model $f$ on task $t$. \textsl{All metrics range from 0 to 100}. Then, we calculate the average score as:
\[
    \textrm{Average Score}(f,\mathcal{A}) = \frac{\sum\limits_{t\in\mathcal{T}} \sum\limits_{s\neq t \in\mathcal{T}} score(s,t,f,\mathcal{A})}{|\mathcal{T}| \times (|\mathcal{T}|-1)}
\]
where $\mathcal{T}$ is the set of tasks.

Additionally, we calculate top-1 score to understand how models and algorithms perform if the best source task is known ahead of time. This score is calculated as the maximum score over source tasks averaged over target tasks. The top-1 score does not consider scores less than the baseline, which is a model trained directly on the target task. Denote the baseline algorithm by $\mathcal{A}_{\mathcal{B}}$ and the baseline score as $score(s,t,f,\mathcal{A}_{\mathcal{B}})$. Formally, the top-1 score is calculated as:
\begin{multline*}
    \textrm{Top-1}(f,\mathcal{A}) =\\
    \frac{\sum\limits_{t\in\mathcal{T}} \max\limits_{s\neq t \in\mathcal{T}} \Big( score(s,t,f,\mathcal{A}_{\mathcal{B}}),score(s, t, f, \mathcal{A})\Big) }{|\mathcal{T}|}
\end{multline*}

\paragraph{Average and Top-1 $\Delta$s}
In addition to raw scores, we also calculate score differences to measure how much a source task benefits a target task. The average $\Delta$ describes how much benefit the model saw in the aggregate over all source tasks, while the top-1 $\Delta$ considers only the best source. Score $\Delta$s are calculated with respect to the baseline score as:
\begin{multline*}
    \Delta(s,t,f,\mathcal{A})=\\score(s,t,f,\mathcal{A})-score(s,t,f,\mathcal{A}_{\mathcal{B}})
\end{multline*}

and the average $\Delta$ is calculated as:
\[
    \textrm{Average }\Delta(f,\mathcal{A})=
    \frac{\sum\limits_{t\in\mathcal{T}} \sum\limits_{s\neq t \in\mathcal{T}}\Delta(s,t,f,\mathcal{A})}
    {|\mathcal{T}| \times (|\mathcal{T}|-1)}
\]
Additionally, we calculate the top-1 $\Delta$ as the maximum positive score difference over source tasks averaged over target tasks:
\[
    \textrm{Top-1 }\Delta(f,\mathcal{A}) =
    \frac{\sum\limits_{t\in\mathcal{T}} \max\limits_{s\neq t \in\mathcal{T}} \Big( 0,\Delta(s,t,f,\mathcal{A})\Big) }{|\mathcal{T}|}
\]

\section{Task Transfer Algorithms}
\label{sec:methods}

In this work, we consider three commonly used task transfer methods: Pre-train/Fine-tune, Multitask, Multitask/Fine-tune. We apply these methods with cross-entropy loss to further optimize pretrained language models on FETA.

\paragraph{Pre-train/Fine-tune} Commonly used in NLP today, the pre-train/fine-tune algorithm consists of two stages of training \cite{transfer_91}.
First, the model is trained on the source task $\mathcal{T}_{S}$, optimizing Eq~\ref{eq:pretrain-on-source}, followed by a separate stage of training on the target task $\mathcal{T}_{T}$, optimizing Eq~\ref{eq:finetune-on-target}:
\begin{equation}\label{eq:pretrain-on-source}
    \mathcal{L}_{S}=\mathop{-\mathbb{E}}_{(x,y_{s})\sim \{X,\mathcal{Y}_{S}\}}\big[\log p(y_{s}|x)\big]
\end{equation}
\begin{equation}\label{eq:finetune-on-target}
    \mathcal{L}_{T}=\mathop{-\mathbb{E}}_{(x,y_{t})\sim \{X,\mathcal{Y}_{T}\}}\big[\log p(y_{t}|x)\big]
\end{equation}

\begin{table*}
    \centering
    \small
    \begin{tabular}{l l | c c c c | c c c c}
        &  & \multicolumn{4}{c|}{\textbf{DailyDialog}} & \multicolumn{4}{c}{\textbf{Friends}} \\
          & \multirow{2}{*}{\begin{tabular}{c}\quad\textbf{Transfer}\\\quad\textbf{Algorithm}\end{tabular}} &  
         \multicolumn{2}{c}{Average} & \multicolumn{2}{c|}{Top-1 Source} & \multicolumn{2}{c}{Average} & \multicolumn{2}{c}{Top-1 Source} \\
         \textbf{Model} & & 
         Score ($\sigma$) & $\Delta$ & Score & $\Delta$ & Score ($\sigma$) & $\Delta$ & Score & $\Delta$ \\
         \hline
         \multirow{3}{*}{BERT} & Pre-train/Fine-tune & 50.61 (0.24) & -0.93 & 52.22 & +0.68 & 42.39 (0.30) & -0.89 & 44.36 & +1.08\\
         & Multitask & 50.95 (0.24) & -0.59 & 52.40 & +0.86 & 42.88 (0.29) & -0.40 & 45.14 & +1.86 \\
         & Multitask/Fine-tune & \underline{\textbf{51.40}} (0.25) & -\underline{0.15} & \underline{52.76} & +\underline{1.22} & \underline{\textbf{44.69}} (0.28) & +\underline{\textbf{1.41}} & \underline{\textbf{46.00}} & +\underline{\textbf{2.72}}\\
         \hline
         \multirow{3}{*}{GPT-2} & Pre-train/Fine-tune & 39.80 (0.25) & -1.28 & 42.19 & +1.11 & 32.66 (0.18) & -0.64 & 34.34 & +1.04 \\
         & Multitask & 40.21 (0.24) & -0.86 & 41.77 & +0.69 & 33.10 (0.16) & -0.20 & 34.83 & +1.53\\
         & Multitask/Fine-tune & \underline{41.15} (0.23) & +\underline{0.07} & \underline{42.76} & +\underline{1.68} & \underline{34.62} (0.15) & +\underline{1.32} & \underline{35.86} & +\underline{2.56}\\
         \hline
         \multirow{3}{*}{T5} & Pre-train/Fine-tune & 49.92 (0.37) & +0.19 & \underline{\textbf{53.04}} & +\underline{\textbf{3.31}} & 41.73 (0.19) & -1.10 & 43.52 & +0.69\\
         & Multitask & 49.49 (0.42) & -0.24 & 52.98 & +3.25 & 40.42 (0.20) & -2.40 & 43.33 & +0.51\\
         & Multitask/Fine-tune & \underline{50.29} (0.36) & +\underline{\textbf{0.56}} & 52.85 & +3.12 & \underline{42.29} (0.17) & -\underline{0.53} & \underline{43.87} & +\underline{1.05}\\
    \end{tabular}
    \caption{\textbf{Average and Top-1 Source task transfer scores.} Average scores and $\Delta$s aggregate scores over all source tasks, compared with Top-1 scores and $\Delta$s which are calculated with scores from the highest performing source task. $\Delta$s are the difference from the baseline score without task transfer. Highest values for each model are underlined, highest values across all models are bolded.}
    \label{tab:overall_transfer}
\end{table*}

\paragraph{Multitask} In this algorithm, there is only a single stage of multitask training \cite{multitask_94}. 
Formally, the training is conducted on both the source and target task by optimizing Eq~\ref{eq:multitask}:
\begin{equation}\label{eq:multitask}
    \begin{split}
        & \mathcal{L}_{S,T} = \\
        & \mathop{-\mathbb{E}}_{(x,y_{s},y_{t})\sim \{X,\mathcal{Y}_{S},\mathcal{Y}_{T}\}}\big[ \log p(y_{s}|x)+\log p(y_{t}|x)\big]
    \end{split}
\end{equation}

\paragraph{Multitask/Fine-tune} 
This algorithm combines the previous algorithms in two stages. In the first stage, the source and target task are optimized jointly, as in Eq~\ref{eq:multitask}. Then, the second stage trains using only the target task, as in Eq~\ref{eq:finetune-on-target}.

Even though model selection in multitasking is generally done w.r.t. multiple source and target tasks \cite{multitask_94}, we modify the setting to validate a model on a single target task at a time. This allows hyperparameter search and early stopping to be controlled by the desired target task.
\section{Experiment Setup}
\label{sec:experiments}

To study task transfer on \feta, we run extensive experimentation. We utilize three task transfer algorithms: pre-train/fine-tune, multitask, and multitask/fine-tune, as described in Section \ref{sec:methods}. To draw broad conclusions about the performance of each learning algorithm, we utilize pretrained language models with three different architectures: encoder-only (BERT) \cite{devlin-etal-2019-bert}, decoder-only (GPT-2) \cite{radford2019language}, and encoder-decoder (T5) \cite{raffel_t5}. Implementation details, including hyperparameters and prompts, can be found in Appendix \ref{sec:implementation_details}.

A complete experiment for a single target task, $\mathcal{T}$, is as follows: First, we directly fine-tune on $\mathcal{T}$ to get the baseline score. Then, for each source task, $\mathcal{S}$, we take the model pre-trained on $\mathcal{S}$ and fine-tune on $\mathcal{T}$. Next, we jointly train on $\mathcal{S}\textrm{ and }\mathcal{T}$ together. Finally, we fine-tune the jointly trained model on $\mathcal{T}$.

\feta\;datasets have 10 and 7 tasks, giving $90+42=132$ unique source-target task pairs. Our experiments include three learning algorithms, three models, and we run each experiment with 5 random seeds. In total, we run $132\times 3\times 3\times 5=5940$ transfer experiments, and $17\times 3\times 5=255$ baseline experiments leading to 6195 trained models.

\begin{figure*}
    \centering
    \includegraphics[width=\textwidth]{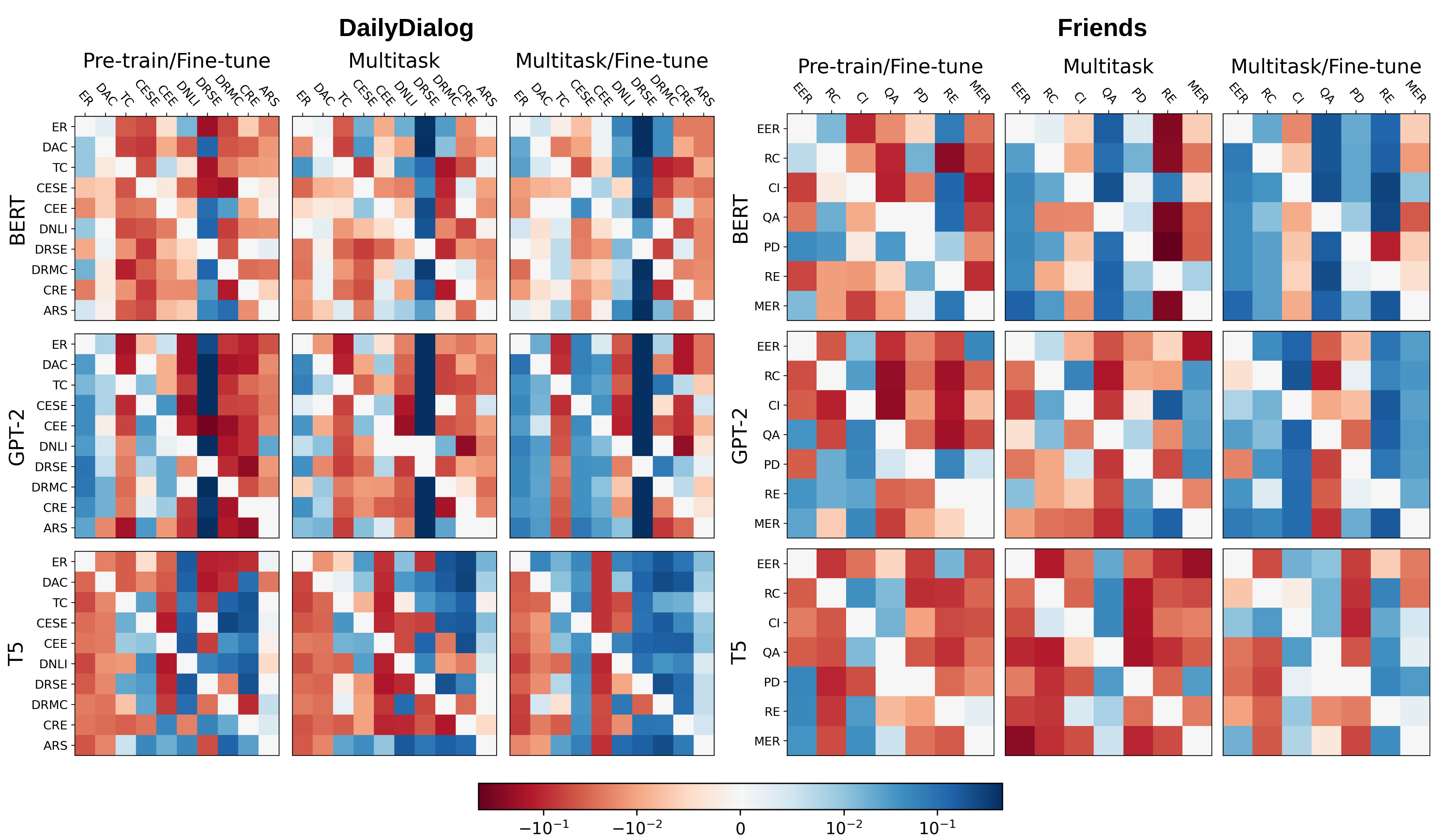}
    \caption{\textbf{Relative improvement of transfer over fine-tuned baselines}. Rows are source tasks and columns are target tasks. Diagonal cells are baseline scores. Looking at an individual column can demonstrate best source tasks for that target. Looking at rows can determine which source task works well across multiple targets.}
    
    \label{fig:main_heatmaps}
\end{figure*}

In addition to the single-source setting described above, we also consider a subset of tasks to study in the multi-source setting, where multiple tasks are simultaneously used as source tasks to transfer to a single target task (\ref{sec:multi_source}). For our experiments, we select two target tasks from each dataset that benefit the most from task transfer, and we use the three source tasks that transferred best onto those targets.
\section{Results and Analysis}
\label{sec:analysis}

\subsection{Single-Source Setting}
Table \ref{tab:overall_transfer} shows the results for all three models and algorithms, and we use this table to understand general trends. Figure \ref{fig:main_heatmaps} shows the relative improvement of a source task for each target task, demonstrating trends across tasks.

\paragraph{Aggregate Performance} We find that, on average, Friends tasks get scores between 7-8 points less than DailyDialog, likely due to the greater number of speakers and utterance length of Friends.
We find that GPT-2 lags behind the raw scores of BERT and T5 by $\sim$10 points. This is expected as autoregressive decoder models are not designed with classification in mind.
We find that the largest average $\Delta$ is 1.4, leaving room for improvement in task transfer on \feta.

Furthermore, we are interested in knowing: how much we would gain by using the best source task vs. a random source task. We calculate the differences between average $\Delta$ and top-1 $\Delta$ and find the mean difference to be $\sim$1.6 and the largest difference to be $\sim$3.5, motivating a further understanding of which source tasks transfer best to target tasks.

\paragraph{Performance Across Learning Algorithms}
We average scores across both datasets and find that pre-train/fine-tune gets an average score of 42.85, multitask 42.84, and multitask/fine-tune 44.07.
Table \ref{tab:overall_transfer} shows that multitask/fine-tune achieves the best average score for all models and datasets, and indeed its average score is a 2.8\% improvement over the other algorithms. However, aggregate scores obscure some interesting nuances.

\paragraph{Do Trends Vary Across Models?}
Previous studies on task transfer have focused on a single model \cite{pruksachatkun-etal-2020-intermediate,Lourie2021UNICORNOR,aribandi2022ext}, but we find that trends
vary depending on the model.
For example, we find results similar to \citet{Lourie2021UNICORNOR}, namely, that fine-tuning on the target task always benefits the T5 model. However, we discover that this does not hold for BERT and GPT-2, which achieve better scores from multitasking than pre-train/fine-tune.

Furthermore, Figure \ref{fig:main_heatmaps} shows that trends on individual tasks also vary depending on the model. For example, T5 positively transferred knowledge to question answering with all learning algorithms and from most source tasks, while GPT-2 had a negative transfer from all algorithms and sources.


For \textit{nearly all} dimensions of analysis (e.g., sample sizes, learning algorithm), we find different trends between models. We \textit{strongly suggest that future research be performed on multiple models} before attempting to draw broad conclusions on transfer learning.

\paragraph{Multitask/Fine-tune As Regularization} We find that T5's top-1 score and $\Delta$ on DailyDialog are highest for pre-train/fine-tune, but the average score and $\Delta$ are highest for multitask/fine-tune. To understand why this occurred, we find the bottom-1 scores for T5 on DailyDialog: 46.78, 46.69, and 48.26 for pre-train/fine-tune, multitask, and multitask/fine-tune algorithms, confirming that multitask/fine-tune does achieve the best worst-case performance.
Moreover, we find that for all datasets and models, multitask/fine-tune does achieve the best worst-case performance. In fact, for GPT-2 on Friends, utilizing the bottom-1 source tasks still lead to a 0.74\% improvement over the baseline.

\begin{figure}[t]
    \centering
    \includegraphics[width=\columnwidth]{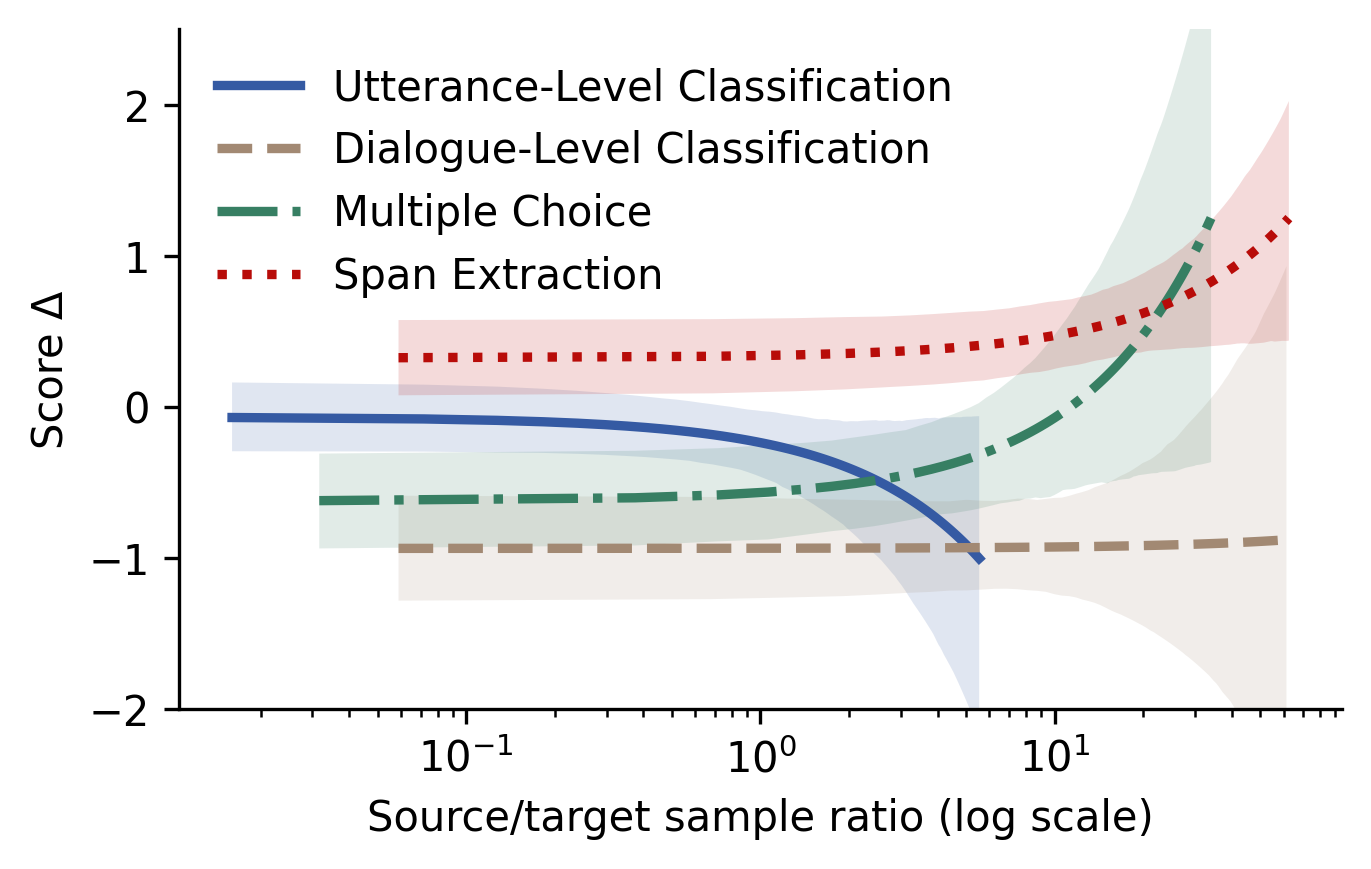}
    \caption{\textbf{Score $\Delta$ by target task type}. Lines show the average score $\Delta$ when the target task is of the specified task type, computed as a best-fit linear interpolation of the data with a 95\% confidence interval. The number of samples for an individual task are fixed, but source/target ratios vary depending on which task pair is used.}
    \label{fig:task_type}
\end{figure}

\paragraph{Do All Task Types Benefit Equally?}
We find that \textit{span extraction tasks gain the most as target tasks}, shown in Figure \ref{fig:task_type} to benefit at all source-to-target sample ratios. Multiple choice tasks also stand to gain from task transfer, but we find that only occurs at a 10:1 ratio of source-target samples. This gain is likely due to the high-level language understanding required by both tasks.

Additionally, we find that utterance-level classification tasks decrease in score $\Delta$ at increasing source-to-target sample ratios. This is possibly due to models overfitting to specific tasks and a catastrophic forgetting of general skills learned during their large-scale pre-training.

\begin{figure}[t]
    \centering
    \includegraphics[width=\columnwidth]{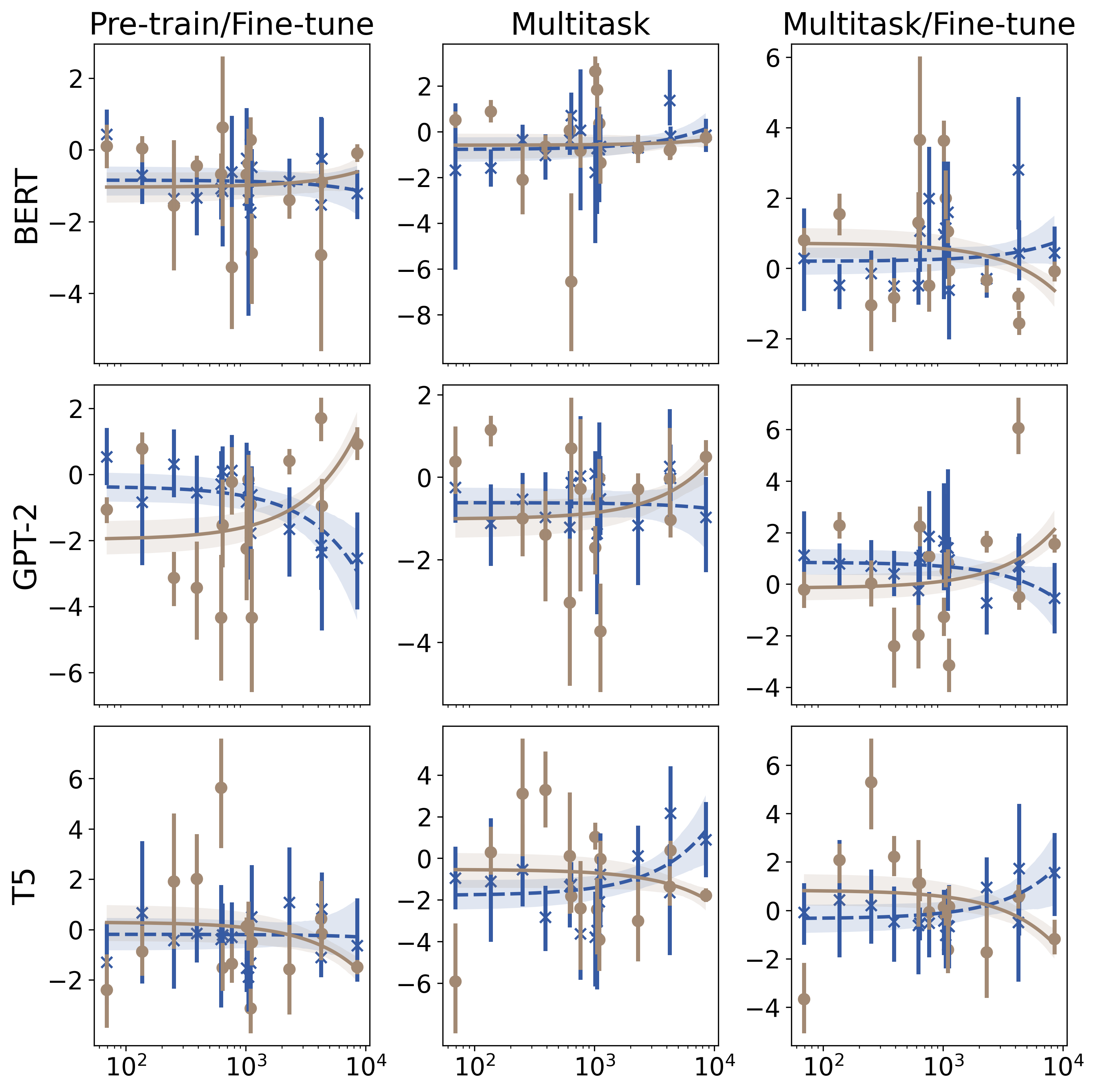}
    \caption{\textbf{Score $\Delta$ by sample count}. Sample count is on the x-axis (log scale) and score $\Delta$ is on the y-axis. The \textcolor{azure}{blue dotted line} represents the average transfer $\Delta$ from a source task to all target tasks. The \textcolor{donkey_brown}{brown line} represents the average transfer $\Delta$ to a target task from all sources. Trend lines are a linear best-fit on the data with a 95\% confidence interval. The number of samples for an individual task are fixed, but source/target ratios vary depending on which task pair is used.}
    \label{fig:score_delta_by_sample_count}
\end{figure}

\paragraph{Do All Task Types Give Equal Benefit?}
We find that \textit{multiple-choice tasks give the greatest benefit as source tasks}, especially when the ratio of source-to-target samples is low, as shown in Figure \ref{fig:score_delta_by_source} in the Appendix. Additionally, we find that at a ratio of 10:1 source-target samples, dialogue-level classification benefits downstream tasks, but utterance-level classification requires a ratio of 100:1.

\paragraph{How Do Sample Sizes Affect Transfer?}
Figure \ref{fig:score_delta_by_sample_count} shows that, interestingly, GPT-2 and T5 have opposite trends in relation to sample size.
We find that $\Delta$s for GPT-2 increase with high target samples and decrease with high source samples. This suggests that GPT-2 may be overfitting to the source task and performs better with resource-rich target tasks. We find that T5 $\Delta$s decrease as target-task samples increase, \textit{suggesting that T5 is more sample efficient} than both GPT-2 and BERT.

\subsection{Multi-Source Setting}
\label{sec:multi_source}
For multi-source transfer we select the two target tasks from each dataset with the best score differences from the single-source setting, shown in Figures \ref{fig:dd_aggregate_heatmap} and \ref{fig:friends_aggregate_heatmap} in the Appendix. We find those four tasks to be Dialogue Reasoning Span Extraction (DRSE), Dialogue-Level NLI (DNLI), Character Identification (CI), and Question Answering (QA). For each of these target tasks, we select the top-3 best source tasks, shown in
Table \ref{tab:multisource_full} of the Appendix
. Learning in this setting is similar to single-source, except we now simultaneously optimize the loss for multiple source tasks.
Table \ref{tab:multisource_results} shows the multi-source results compared with the average score of the top-3 source tasks from the single-source setting. Full results, including score $\Delta$s from the single-source baselines, average top-3 score $\Delta$s, and multi-source score $\Delta$s are in Table \ref{tab:multisource_full} of the Appendix.

\begin{table}[t]
    \centering
    \small
    \begin{tabular}{l l | c | c | c | c}
        \multicolumn{2}{r|}{\textbf{Target}} & \textbf{DRSE} & \textbf{DNLI} & \textbf{CI} & \textbf{QA}\\
        \hline
        \parbox[t]{2mm}{\multirow{3}{*}{\rotatebox[origin=c]{90}{\textbf{BERT}}}}
        & P/F & -1.18 & +1.37& -2.11 & -0.99\\
        & M & +2.77 & +1.57 & -0.54 & -1.14\\
        & M/F & +1.61 & +2.28 & -0.34 & -0.55\\
        \hline
        \parbox[t]{2mm}{\multirow{3}{*}{\rotatebox[origin=c]{90}{\textbf{GPT-2}}}}
        & P/F &  +0.40 & +0.16 & +4.25 & -3.90\\
        & M & +0.78 & +0.98 & +1.28 & -2.46\\
        & M/F & +0.73 & -0.09 & +0.00 & -0.95\\
        \hline
        \parbox[t]{2mm}{\multirow{3}{*}{\rotatebox[origin=c]{90}{\textbf{T5}}}}
         & P/F &  +0.60 & +1.95 & -0.79 & +0.48\\
        & M & -1.08 & -0.96 & -1.49 & +0.08\\
        & M/F & -1.22 & -1.20 & -0.24 & -0.22
    \end{tabular}
    \caption{\textbf{Multi-source score $\Delta$s from the average score of the top-3 source tasks.} Full results, including score $\Delta$s from the fine-tuned baseline are in Table \ref{tab:multisource_full}.}
    \label{tab:multisource_results}
\end{table}

\paragraph{Does Multi-source Improve Over Single-source?}We expect that by utilizing the top-3 source tasks from the single-source setting, the multi-source setting will improve performance for all models and algorithms, but find results to the contrary. We find that 6/9 multi-source algorithms outperform their average top-3 single-source counterparts in DRSE, 6/9 for DNLI, 3/9 for CI, and only 2/9 for QA, showing that naively combining source tasks is not always beneficial.
The impressive result for DRSE follows our original intuition, given that there is an almost unanimous benefit from all source tasks, shown in Figure \ref{fig:main_heatmaps}. Similarly, we find that \textit{multi-source performance on CI also correlates with the performance of individual source tasks}. We find that in the single-source setting GPT-2 is the only model that improves with any source task, and indeed GPT-2 sees benefits from multi-source training on all algorithms.

\paragraph{Which Models Benefit From Multi-Source?}
Table \ref{tab:multisource_full} shows that GPT-2 improves in 8/12 experiments over its average top-3 single-source counterparts, but BERT only 5/12 and T5 in only 4/12 experiments. 
It is counter-intuitive that T5 should perform the worst as we expect that it has a higher capacity for learning due to twice the model size. On the other hand, the additional parameters may be causing T5 to overfit on training data in the few-sample setting.
\section{Conclusion}
We introduce \feta, a comprehensive benchmark for evaluating language models and task transfer learning algorithms in open-domain dialogue with few samples. Through extensive experimentation, we find new and non-intuitive insights on the mechanisms of transfer learning. In particular, we find that most trends are model-specific, and we strongly encourage researchers to consider multiple model architectures before attempting to draw broad conclusions on transfer learning.
It is our hope that \feta\;enables further research not only in task transfer, but also in other learning settings, and in the generalizability and efficiency of model architectures and pre-training datasets.

\section*{Limitations}

A concern regarding any work that includes large-scale experiments with large language models is the energy consumption and environmental impact, the current work included.
While there is a cost to running these experiments, the goal of this work is to improve sample efficiency in the future and we hope that the benefits in future energy saved will outweigh the up-front costs of discovering efficient methods.

Another concern of a large-scale benchmark is that of accessibility. A benchmark requiring too many resources will limit those who can reasonably compete. For this reason and others, in addition to our large-scale benchmark we also include a smaller multi-source setting which requires only 4 experiments to be run for a single model and algorithm, rather than 132 in the single-source setting. We believe this smaller setting will maintain the ability to extract high-quality insights on task transfer, yet allow for increased community access and reduce the carbon footprint of this benchmark.

While we do control for domain adaptation in our experiments on task transfer, there are some aspects that we cannot control. For example, each model has done language model pre-training with a different corpus.
BERT was trained on English Wikipedia and BookCorpus \cite{Zhu_2015_ICCV}, GPT-2 was trained on a WebText \cite{radford2019language}, and T5 was trained on C4 \cite{raffel_t5}.
This difference likely affects model performance on the dialogue tasks in \feta.

Additionally, we cannot exhaustively test every language model, but still try to provide enough variety in order to draw broad conclusions on task transfer. For example, we don't run any experiments on language models pre-trained in the dialogue domain or language models larger than base-sized. We expect that both of these changes would improve raw performance on \feta. More importantly though, it is unclear whether either of these changes would lead to improved task-transfer performance (average and top-1 $\Delta$s) and we leave this exploration for future work.

Furthermore, we cannot exhaustively test all learning algorithms. For example, \citet{wang-etal-2020-negative} propose a transfer learning method that minimizes negative task interference via meta-learning for multilingual models, \citet{albalak-etal-2022-rex} propose a policy-guided algorithm for task transfer in low-data settings, and \citet{gradient_surgery} propose an optimization algorithm that mitigates gradient interference for reinforcement learning agents.

Finally, we stress the importance of \textit{intra-dataset} task transfer in this work. However, this limits the number of pre-annotated tasks that are available, and there are certainly some tasks which we were not able to accomodate in \feta.
\section*{Acknowledgements}

The authors would like to thank William Cohen and Tania Bedrax-Weiss for their valuable insights and constructive feedback about this work. This material is based on work that is partially funded by an unrestricted gift from Google. This work was supported by the National Science Foundation award \#2048122. The views expressed are those of the author and do not reflect the official policy or position of the US government. Finally, we thank the Robert N. Noyce Trust for their generous gift to the University of California via the Noyce Initiative.

\bibliography{anthology,references}
\bibliographystyle{acl_natbib}

\appendix

\section{Dataset Details}
\label{sec:dataset_details}
\begin{figure}[t]
    \centering
    \includegraphics[width=\columnwidth]{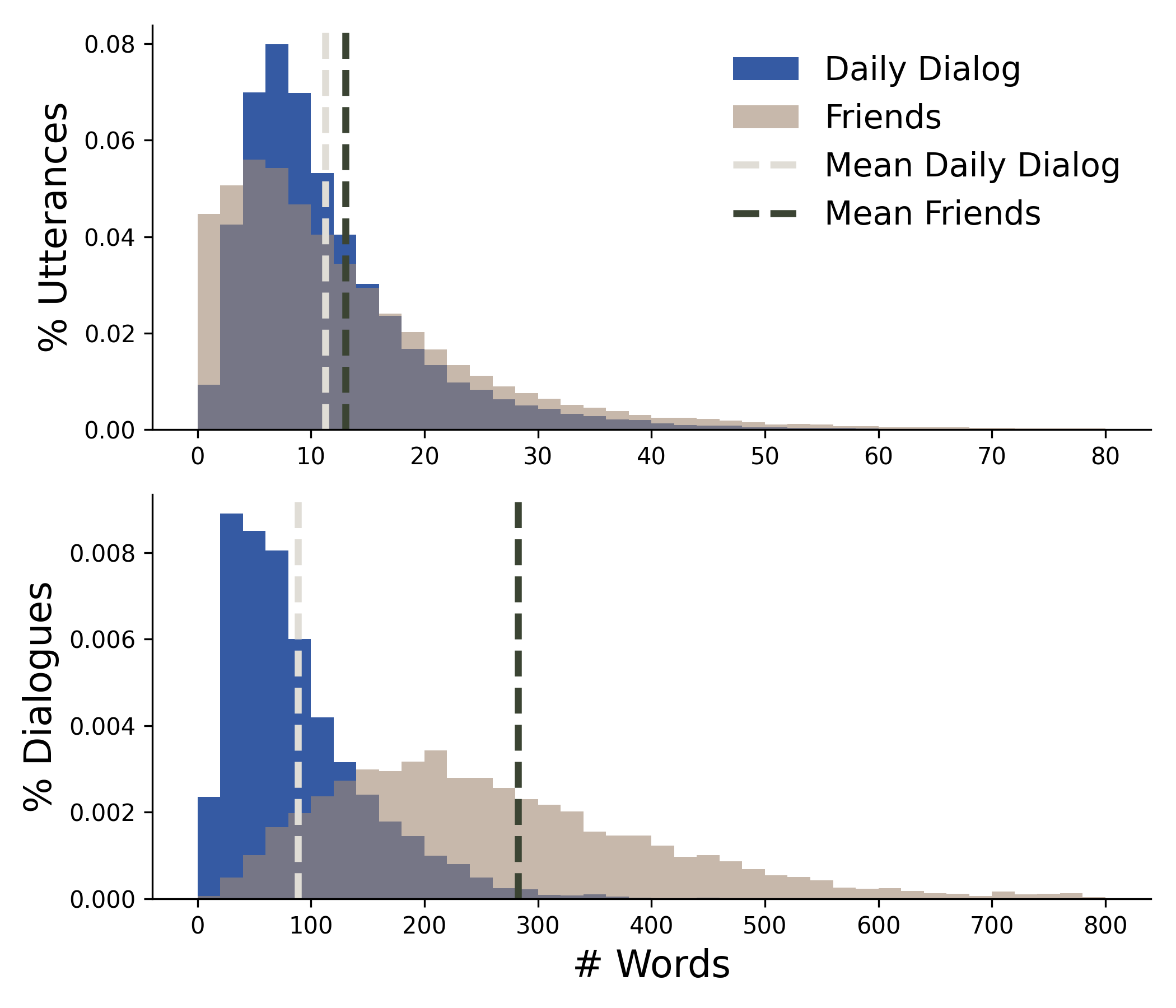}
    \caption{\textbf{Utterance and dialogue length distributions in \feta.}}
    \label{fig:dialog_lens}
\end{figure}

\subsection{DailyDialog}
\paragraph{DailyDialog} Along with the dialogues, \citet{li-etal-2017-dailydialog} provide annotations for \textbf{emotion recognition}, \textbf{dialogue act classification}, and \textbf{topic classification}.

\paragraph{RECCON} \citet{reccon} introduce the task of recognizing emotion causes in conversation and provide annotations for two subtasks: \textbf{causal emotion span extraction} and \textbf{causal emotion entailment}. Recognizing the cause behind emotions is an important aspect of developing conversational agents that can respond appropriately and these tasks test that ability. Both tasks assume that the emotion of an utterance is already known and require a model to identify the evidence or cause of the given emotion. In causal emotion span extraction, the model is given input as "The target utterance is <U\textsubscript{t}>. The evidence utterance is <U\textsubscript{e}>. What is the causal span from evidence in the context that is relevant to the target utterance's emotion <E\textsubscript{t}>?".
On the other hand, if the conversation history up to utterance U\textsubscript{t} is H(U\textsubscript{t}), then the task of causal emotion entailment is to classify the triple (U\textsubscript{t},U\textsubscript{e},H(U\textsubscript{t})) as entailment or not entailment. In this case, entailment means that the emotion expressed in the target utterance, U\textsubscript{t}, is caused by the evidence utterance, U\textsubscript{e}.

\paragraph{CIDER} \citet{ghosal-etal-2021-cider} provide annotations for four tasks designed to explore commonsense inference and reasoning in dialogue: \textbf{dialogue-level natural language inference} (DNLI), \textbf{dialogue reasoning span extraction}, \textbf{dialogue reasoning multiple choice}, and \textbf{commonsense relation extraction}. These tasks are created by annotating knowledge triplets on 31 relations that are either explicitly stated in the dialogue or that require commonsense reasoning using contextual information. In DNLI, the task is to determine whether a triplet is true or false given the dialogue. Given a knowledge triplet as <head, relation, tail>, the span extraction task is formulated as identifying the tail when given the head, relation, and dialogue for context. The multiple choice task is motivated by the SWAG commonsense inference task \cite{zellers2018swagaf}, given a head, relation, and conversation as context, the goal is to predict the tail of the relation from 4 possible choices. Finally, commonsense relation extraction is formulated as usual relation extraction tasks; given the head, tail, and conversation as context, the goal is to predict the correct relation out of 31 options.

\paragraph{DailyDialog++} \citet{dailydialog_plusplus} present the DailyDialog++ dataset, where they aim to improve evaluation of response generation. They do so by collecting five relevant responses and five adversarially crafted irrelevant responses for each dialogue in their dataset, and we recycle their data for a new task called \textbf{adversarial response selection}. Adversarial response selection is formulated as a multiple choice selection between a correct response, a randomly selected negative response, and an adversarial negative response.

\subsection{Friends}
\paragraph{EmoryNLP} \citet{chen-choi-2016-character} and \citet{zhou-choi-2018-exist} provide annotations for \textbf{character identification}, a subtask of entity linking, where entity mentions in an utterance need to be matched to their correct entity. For this task there are seven possible entities: the six main characters and an "other" entity.

\citet{zahiri2018emotion} provide annotations on \textbf{emotion recognition}, with the 7 fine-grained emotions from the Feeling Wheel \cite{feeling_wheel}.

\citet{ma-etal-2018-challenging} present annotations for a subtask of \textbf{reading comprehension}, called passage completion. In passage completion, given a dialogue and factual statement about the dialogue where character mentions are removed, the task is to fill in the blanks with the correct character from the dialogue. This task is similar to a multiple choice task because entity choices are presented to the model, but because there are varying number of options in each dialogue, it is formulated as a span extraction that is evaluated based on accuracy.

\citet{yang-choi-2019-friendsqa} introduce annotations for \textbf{question answering}. The answers to question-answer pairs can either be a speaker name or exist as a span within the dialogue, and multiple spans may be correct.

\citet{jiang2020automatic} present the \textbf{personality detection} task by annotating speakers with five traits: agreeableness, conscientiousness, extraversion, openness, and neuroticism. The goal of the task is to correctly identify whether a given character from a dialogue either has or does not have each of the five traits.

\paragraph{DialogRE} \citet{yu2020dialogue} introduce a \textbf{relation extraction} dataset annotated with 36 different relations. Their dataset anonymizes speakers which allows for an entity linking relation called "per:alternative\_name". However, our version of the Friends dataset is named and so we remove this relation from our data. This task is similar to the relation extraction from DailyDialog, however the relations in DailyDialog are commonsense relations, and the relations in Friends are focused on information about entities.

\paragraph{MELD} \citet{poria2019meld} provide additional annotations for \textbf{emotion recognition}, with only 22.2\% dialogue overlap with \citet{zahiri2018emotion}'s dialogues. Additionally, while both use 7 total emotions, \citet{poria2019meld} use 2 different emotions from \citet{zahiri2018emotion}.

\section{Implementation Details}
\label{sec:implementation_details}
For our experiments, we use the pretrained model implementations from the HuggingFace Transformers library \cite{wolf-etal-2020-transformers}, where the bert-base-uncased model has 110M parameters, GPT-2 has 124M parameters, and T5-base has 223M parameters. We use the Adam optimizer \cite{DBLP:journals/corr/KingmaB14} with a batch size of 60 and run a learning rate sweep across \{3$\times$10\textsuperscript{-6}, 1$\times$10\textsuperscript{-5},3$\times$10\textsuperscript{-5},1$\times$10\textsuperscript{-4}\} during the pre-training phase, finding that 3$\times$10\textsuperscript{-5} worked well across all models. In all experiments we utilize validation-based best model selection, and train models for 30 epochs on DailyDialog tasks and 20 epochs on Friends tasks.

\begin{table*}[t]
    \centering
    \begin{tabular}{l|l}
         \textbf{Task}& \textbf{Prompt} \\
         \hline
         Emotion Recognition & 
         emotion: \\
         \hline
         Dialogue Act Classification & dialogue act: \\
         \hline
         Topic Classification & topic:\\
         \hline
         Causal Emotion Span Extraction & 
         question: <question> answer:
         \\
         \hline
         Causal Emotion Entailment &
         context: <premise> causal emotion entailment: <hypothesis>
         \\
         \hline
         Dialogue-level NLI & context: <premise> entailment: <hypothesis>
         \\
         \hline
         Dialogue Reasoning Span Extraction & question: <question> answer:
         \\
         \hline
         Dialogue Reasoning Multiple Choice &
         question: <question> <options> The correct option is 
        \\
         \hline
         Commonsense Relation Extraction & The relation between <head> and <tail> is 
         \\
         \hline
         Adversarial Response Selection & question: <question> <options> The correct option is
         \\
         \hline
    \end{tabular}
    \caption{\textbf{Prompts for \feta-DailyDialog tasks}. All prompts start with "context: <context>", but we leave this out due to repetitiveness and space.}
    \label{tab:DD_prompts}
\end{table*}

\begin{table*}[t]
    \centering
    \begin{tabular}{l|l}
         \textbf{Task}& \textbf{Prompt} \\
         \hline
         Emotion Recognition (Emory) & emotion:\\
         \hline
         Reading Comprehension & question: <question> out of <entities> [PLACEHOLDER] is\\
         \hline
         Character Identification & out of <options>, <mention> in the phrase <phrase> refers to \\
         \hline
         Question Answering & question: <question> answer:\\
         \hline
         Personality Detection & <entity> is <characteristic>\\
         \hline
         Relation Extraction & <head> has the following relations with <tail>\\
         \hline
         Emotion Recognition (MELD) & emotion:\\
         \hline

    \end{tabular}
    \caption{\textbf{Prompts for \feta-Friends tasks}. All prompts start with "context: <context>", but we leave this out due to repetitiveness and space.}
    \label{tab:friends_prompts}
\end{table*}

\pagebreak
\section{Expanded Single-Source Results}
\label{sec:expanded_results}
\begin{figure*}
    \centering
    \includegraphics[width=\textwidth]{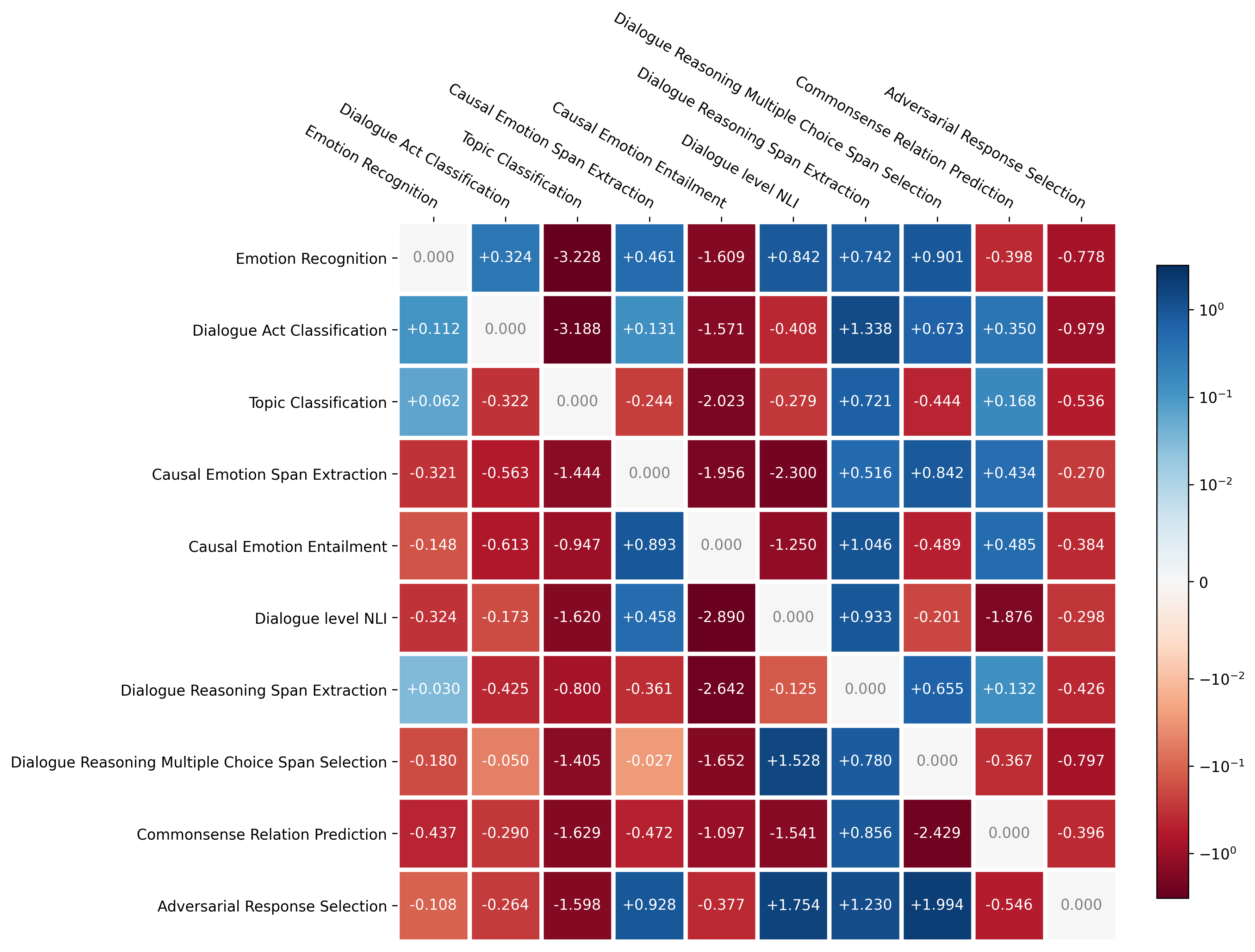}
    \caption{\textbf{Aggregate task transfer performance on DailyDialog.}}
    \label{fig:dd_aggregate_heatmap}
\end{figure*}

\begin{figure*}
    \centering
    \includegraphics[width=\textwidth]{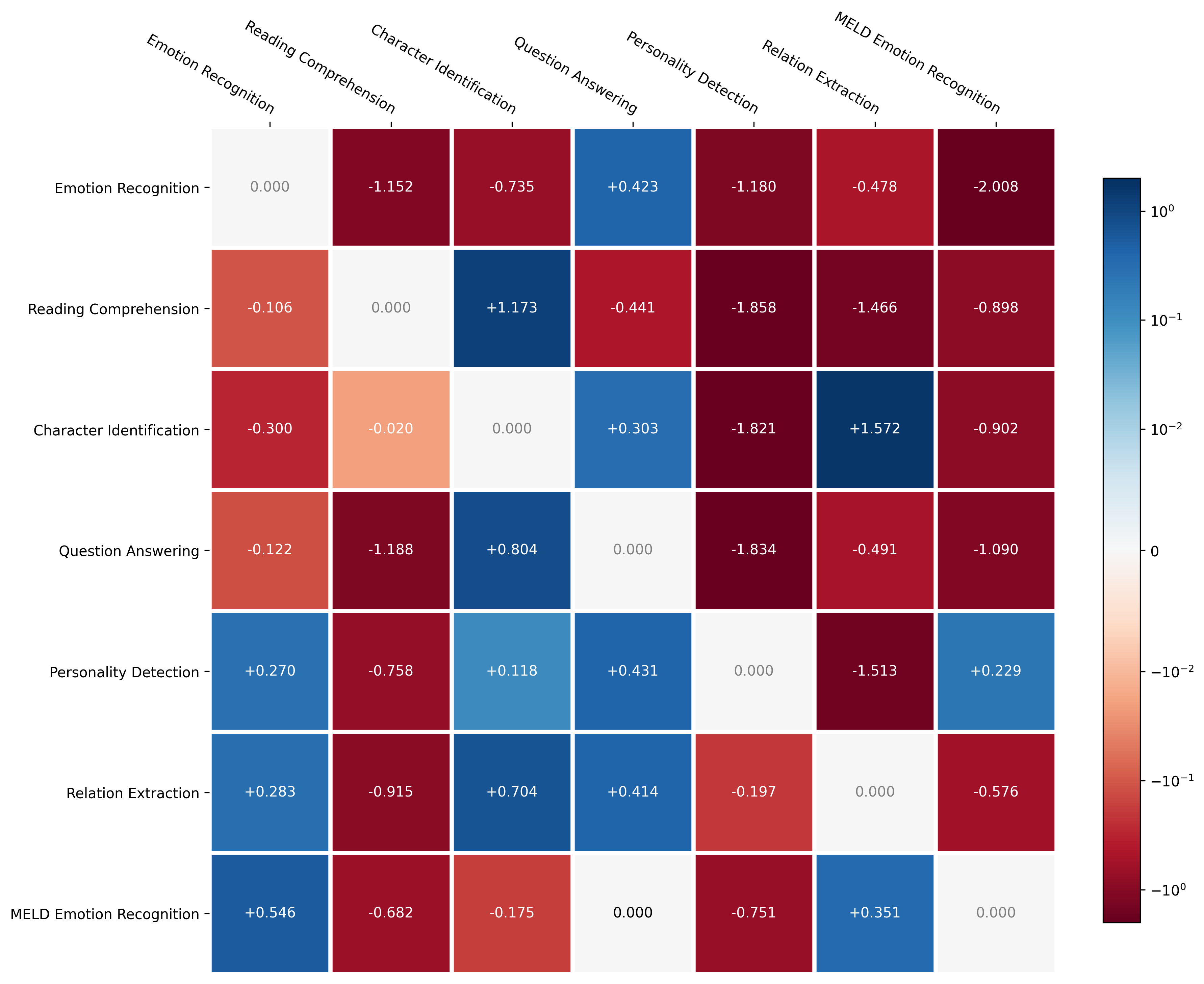}
    \caption{\textbf{Aggregate task transfer performance on Friends.}}
    \label{fig:friends_aggregate_heatmap}
\end{figure*}

\begin{figure}
    \centering
    \includegraphics[width=\columnwidth]{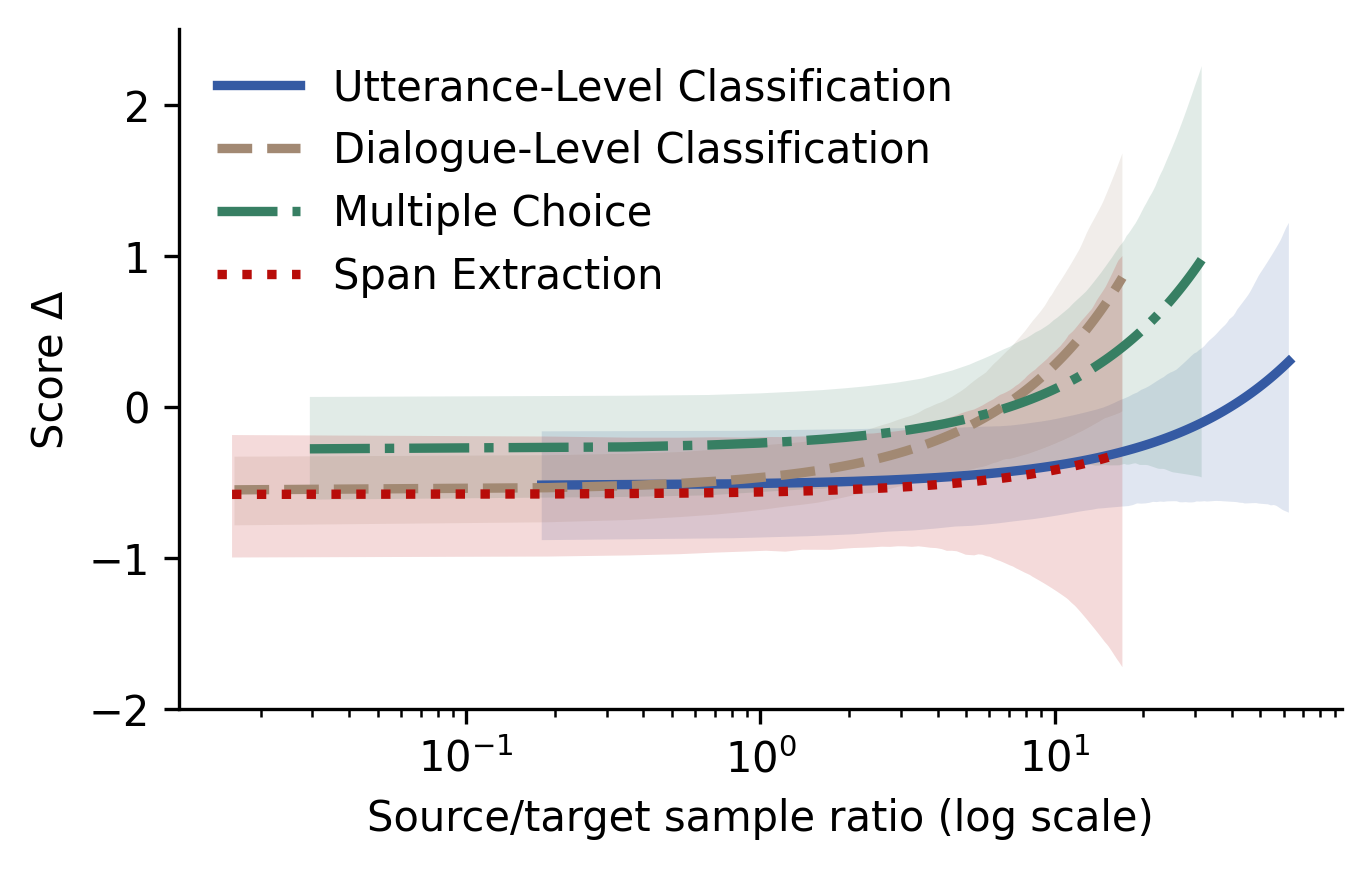}
    \caption{\textbf{Score $\Delta$ by source task type}. The number of samples for an individual task are fixed, but source/target ratios vary depending on which task pair is used..}
    \label{fig:score_delta_by_source}
\end{figure}

\pagebreak

\section{Expanded Multi-Source Results}
\label{sec:expanded_multisource_results}

\begin{table*}
    \centering
    \setlength{\tabcolsep}{2pt}
    \begin{adjustbox}{angle=-90}
    \begin{tabular}{l l | c c c c c| c c c c c | c c c c c | c c c c c}
        \multicolumn{2}{r|}{\textbf{Target}} & \multicolumn{5}{c}{\textbf{DRSE}} & \multicolumn{5}{c}{\textbf{DNLI}} & \multicolumn{5}{c}{\textbf{CI}} & \multicolumn{5}{c}{\textbf{QA}}\\
         &  & & & & \multirow{2}{*}{\begin{tabular}{c}
              \textbf{Top-3}  \\
              \textbf{Av.} 
         \end{tabular}}& \multirow{2}{*}{\begin{tabular}{c}\textbf{Multi-}\\\textbf{Source}\end{tabular}} &
          & & &\multirow{2}{*}{\begin{tabular}{c}
              \textbf{Top-3}  \\
              \textbf{Av.} 
         \end{tabular}}& \multirow{2}{*}{\begin{tabular}{c}\textbf{Multi-}\\\textbf{Source}\end{tabular}} &
          & & &\multirow{2}{*}{\begin{tabular}{c}
              \textbf{Top-3}  \\
              \textbf{Av.} 
         \end{tabular}}& \multirow{2}{*}{\begin{tabular}{c}\textbf{Multi-}\\\textbf{Source}\end{tabular}} &
          & & &\multirow{2}{*}{\begin{tabular}{c}
              \textbf{Top-3}  \\
              \textbf{Av.} 
         \end{tabular}}& \multirow{2}{*}{\begin{tabular}{c}\textbf{Multi-}\\\textbf{Source}\end{tabular}}\\
        & & DAC & ARS & CEE & & & ARS & DRMC & ER & & & RC & QA & RE & & & PD & ER & RE & &  \\
        \hline
        \parbox[t]{2mm}{\multirow{3}{*}{\rotatebox[origin=c]{90}{\textbf{BERT}}}}
        & P/F & 0.46 & 0.17 & 0.43 & 0.35 & -0.83 & -0.35 & -0.39 & 0.88 & 0.05 & \underline{1.48} & -1.21 & -0.76 & -1.15 & -0.16 & -2.27 & 0.58 & -0.28 & -0.08 & 0.07 & -0.92\\
        & M & 1.86 & 0.15 & 0.86 &0.96 & \underline{3.73} & 0.53 & 0.32 & 1.05 & 0.63 & \underline{2.20} & -0.77 & -1.48 & -0.27 & -0.84 & -1.38 & 1.89 & 2.98 & 2.62 & 2.50 & 1.36\\
        & M/F & 2.58 & 2.04 & 1.40 & 2.01 & \underline{3.62} & 2.66 & 0.40 & 3.55 & 2.20 & \underline{4.48} & -0.58 & -0.78 & -0.48 & -0.61 & -0.95 & 3.04 & 3.64 & 4.49 & 3.72 & 3.17\\
        \hline
        \parbox[t]{2mm}{\multirow{3}{*}{\rotatebox[origin=c]{90}{\textbf{GPT-2}}}}
        & P/F & 0.93 & 1.14 & -0.3 & 0.59 & \underline{0.99} & -3.65 & 0.00 & -6.99 & -3.55 & \underline{-3.39} & 1.29 & 2.73 & 1.09 & 1.70 & \underline{5.95} & 0.12 & -1.76 & -0.66 & -0.77 & -4.67 \\
        & M & 1.30 & 1.59  & 0.89 & 1.26 & \underline{2.04} & -0.81 & -1.73 & -0.94 & -1.16 & \underline{-0.18} & 2.70 & -1.03 & -0.26 & 0.47 & \underline{1.75} & -1.59 & -1.00 & -1.14 & -1.24 & -3.70\\
        & M/F & 3.43 & 2.01 & 1.70 & 2.38 & \underline{3.11} & 0.46 & -0.32 & -1.92 & -0.59 & -0.68 & 8.81 & 6.69 & 5.08 & 6.86 & \underline{8.81} & -1.31 & -0.84 & -0.83 & -0.99 & -1.94\\
        \hline
        \parbox[t]{2mm}{\multirow{3}{*}{\rotatebox[origin=c]{90}{\textbf{T5}}}}
         & P/F & -3.08 & -1.08 & -1.48 & -1.88 & \underline{-1.28} & 2.52 & 5.53 & 8.60 & 5.55 & \underline{7.50} & 2.22 & 0.70 & 1.59 & 1.50 & 0.71 & 0.03 & -0.19 & -0.31 & -0.16 & \underline{0.32}\\
        & M & 1.54 & 1.77 & 2.93 & 2.08 & 1.00 & 8.83 & 5.83 & 0.55 & 5.07 & 4.11 & -1.84 & -0.30 & 0.22 & -0.64 & -2.13 & 1.10 & 0.82 & 0.27 & 0.73 & \underline{0.81} \\
        & M/F & 3.00 & 3.30 & 2.99 & 3.10 & 1.88 & 5.59 & 4.10 & 2.78 & 4.16 & 2.96 & -0.06 & 1.46 & 0.52 & 0.64 & 0.40 & 0.02 & 0.42 & -0.63 & -0.06 & -0.28
    \end{tabular}
    \end{adjustbox}
    \caption{\textbf{Results from the multi-source experiment}, where we use the top-3 source tasks in a multi-source task transfer setting. We include individual scores from all 3 top-3 source tasks and include their average score as a comparison. Multi-source experiments that improve over the top-3 average are underlined.}
    \label{tab:multisource_full}
\end{table*}

\end{document}